\documentclass[
]{ceurart}

\usepackage{listings}
\usepackage{float}
\restylefloat{table}  
\usepackage{makecell}
\usepackage{graphicx}
\usepackage{algorithm}
\usepackage{algpseudocode}
\usepackage{amsmath}
\usepackage{placeins}
\usepackage{subfigure}
\begin{document}

\copyrightyear{2025}
\copyrightclause{Copyright for this paper by its authors.
  Use permitted under Creative Commons License Attribution 4.0
  International (CC BY 4.0).}

\conference{MiGA@IJCAI25: International IJCAI Workshop on 3rd Human Behavior Analysis for Emotion Understanding, August 29, 2025, Guangzhou, China.}

\title{Weak to Strong: VLM-Based Pseudo-Labeling as a Weakly Supervised Training Strategy in Multimodal Video-based Hidden Emotion Understanding Tasks}

\tnotemark[1]
\tnotetext[1]{You can use this document as the template for preparing your
  publication. We recommend using the latest version of the ceurart style.}



\author[1]{Yufei Wang}[%
  orcid=0009-0008-6002-3729,
  email=z5536297@ad.unsw.edu.au,
]
\fnmark[1]

\author[2]{Haixu Liu}[%
  orcid=0009-0007-8115-0826,
  email=hliu2490@uni.sydney.edu.au,
]
\fnmark[1]

\author[3]{Tianxiang Xu}[%
  orcid=0000-0002-6121-2432,
  email=xtx_pku@stu.pku.edu.cn,
]
\fnmark[1]

\author[2]{Chuancheng Shi}[%
  orcid=0009-0002-2278-2341,
  email=cshi0459@uni.sydney.edu.au,
]
\fnmark[1]

\author[4]{Hongsheng Xing}[%
  orcid=0000-0002-4901-4863,
  email=starxsky@163.com,
]
\fnmark[1]

\address[1]{The University of New South Wales, Sydney, New South Wales, Australia}
\address[2]{The University of Sydney, Sydney, New South Wales, Australia}

\address[3]{School of Software and Microelectronics, Peking University, Zibo, Shandong, China}

\address[4]{Shandong University of Technology, Beijing, China}

\cortext[1]{Corresponding author.}
\fntext[1]{These authors contributed equally.}

\begin{abstract}
To tackle the automatic recognition of “concealed emotions” in videos, this paper proposes a multimodal weak-supervision framework and achieves state-of-the-art results on the iMiGUE tennis-interview dataset. First, YOLO 11x detects and crops human portraits frame-by-frame, and DINOv2-Base extracts visual features from the cropped regions. Next, by integrating Chain-of-Thought and Reflection prompting (CoT + Reflection), Gemini 2.5 Pro automatically generates pseudo-labels and reasoning texts that serve as weak supervision for downstream models.
Subsequently, OpenPose produces 137-dimensional key-point sequences, augmented with inter-frame offset features; the usual graph neural network backbone is simplified to an MLP to efficiently model the spatiotemporal relationships of the three key-point streams. An ultra-long-sequence Transformer independently encodes both the image and key-point sequences, and their representations are concatenated with BERT-encoded interview transcripts. Each modality is first pre-trained in isolation, then fine-tuned jointly, with pseudo-labeled samples merged into the training set for further gains.
Experiments demonstrate that, despite severe class imbalance, the proposed approach lifts accuracy from under 0.6 in prior work to over 0.69, establishing a new public benchmark. The study also validates that an “MLP-ified” key-point backbone can match—or even surpass—GCN-based counterparts in this task.

\end{abstract}


\maketitle

\section{Introduction}

\subsection{Background and Literature Review}

Video-based hidden emotion understanding refers to the automatic recognition and inference of real emotional states that are deliberately or unintentionally concealed and not directly expressed, by analyzing extremely brief micro-expressions, subtle facial muscle movements, body and gesture movements, or speech and other multimodal temporal signals in videos. Unlike traditional emotion recognition, which focuses on obvious and explicit emotional expressions, hidden emotion recognition focuses on subtle cues such as micro-expressions, micro-movements, and intonation changes, often combining context understanding and multimodal fusion techniques. This task has significant practical value in psychological assessment and clinical intervention, stress and anomaly detection in security and judicial scenarios, and enhancing human-computer interaction experience, and also provides a brand new technical means for basic research in behavioral psychology.
Currently, this task faces challenges such as data scarcity and labeling difficulties, weak and sparse signals, excessively long time series making feature extraction difficult, and insufficient model generalization ability.

We use the iMiGUE dataset\cite{liu2021imigue}, which consists of 359 post-match press conference videos collected from the Australian Open channel on YouTube, covering 72 tennis players from 28 countries and regions, aged 17 to 38, with balanced gender. Each video is labeled for emotion according to the match result (win/loss), as an indirect indicator of emotional expression. To avoid interview audio exposing the interviewee's match result, all videos are muted.
It should be noted that the ratio of positive (win) samples to negative (loss) samples is about 3:1, with positive samples accounting for about 75\%, indicating a significant class imbalance in this dataset.

According to data officially released by iMiGUE, previous work usually uses CNN-based models (e.g., I3D) to process the video modality, or GCN-based models \cite{yu2017spatio} (e.g., STGCN) to process the human pose modality, and then uses LSTM, 1D CNN, or Transformer for time series modeling (e.g., TSM + LSTM). However, the accuracy of the above methods on this task does not exceed 60\%. In recent years, with the development of VLM models, some representative VLM models such as Gemini 2.5 Pro have also acquired the ability for video-based hidden emotion understanding. We tested their performance based on simple and carefully constructed prompts: Gemini 2.5 Pro achieves about 50\% accuracy with simple prompts and up to 64\% accuracy with carefully constructed prompts.

\subsection{Method and Contributions}

Our proposed method adopts YOLO11x to crop rectangular boxes containing human figures in each frame during feature engineering and uses Dinov2 Base \cite{oquab2023dinov2} to encode these boxes, thereby obtaining an image feature time series. We then design a prompt with CoT and reflection techniques to guide the vision-language model to recognize action units in FACS, PGCS, and GACS and analyze the probability of the interviewed tennis player winning or losing based on some prior rules, while outputting the analysis process in text form. We input the video and prompt into the Gemini 2.5 Pro model to convert the video into a text modality and obtain Pseudo-labels for the test set given by Gemini 2.5 Pro. Finally, we use Openpose to identify 137 key points in the human body in each frame, including 25 body skeleton key points, 70 facial key points, and 42 hand key points (21 per hand), and calculate the offset in the horizontal and vertical pixel distances between the current frame key points and the previous 8th, 16th, and 24th frames. These offsets, together with the normalized key point pixel coordinates, form the feature of each key point. Then, for each video, we sample 4000 frames, use graph relations to represent the three sets of key points in each frame, and use a Transformer to process these three ultra-long graph structure time series to model their spatiotemporal relationships, and then concat the features output by the three Transformers and fuse them through a fully connected layer. We use the BERT model to encode the text corresponding to the video and obtain the feature vector of the text modality. At the same time, for the image feature time series previously extracted frame by frame with Dinov2 Base, we ensure the sampling interval is greater than 5 frames, and 800 frames are sampled from each video, and a Transformer is also used to process this ultra-long image feature time series to model the spatiotemporal relationships of the video. Then, we also simply concat the text modality, video modality, and the fused key point modality, and connect them to the output head to form the complete model. For each modality, we pre-train on the current training data, then load the pre-trained backbone network back into the model for fine-tuning. In the above, we obtained pseudo-labels for the test set labeled by Gemini 2.5 Pro. We combined the pseudo-labeled test set and the training set for fine-tuning, obtained the final model, and then obtained the final score after a series of post-processing tricks.

Specifically, our work has the following contributions:
\begin{itemize}
    \item We designed a multimodal model for video-based hidden emotion understanding, which became the SOTA on this task.
    \item We proposed a training paradigm to improve model performance based on weak supervision learning with pseudo-label annotation using a large model.
    \item In comparative experiments, we demonstrated that degrading the graph neural network to an MLP in the backbone network for human key point feature extraction can achieve equal or even better results.
\end{itemize}

\section{EDA}

\subsection{Openpose Keypoint Connection Visualization}

We visualized the keypoints of the body skeleton, face, and hands in the training data according to the keypoint connection relationships provided by Openpose. We found that the recognition of skeleton keypoints is generally accurate, although the lower body is significantly occluded, but the contribution of lower body limb movements to the task can be almost ignored. The recognition of facial keypoints is also generally accurate, so the facial keypoints may contribute more to the task. However, there is a large error in hand keypoint recognition. Although micro-gestures are still important, the difficulty in accurately recognizing hand keypoints may introduce considerable noise during training.

\begin{figure}[h]
	\centering
	\includegraphics[width=1\linewidth]{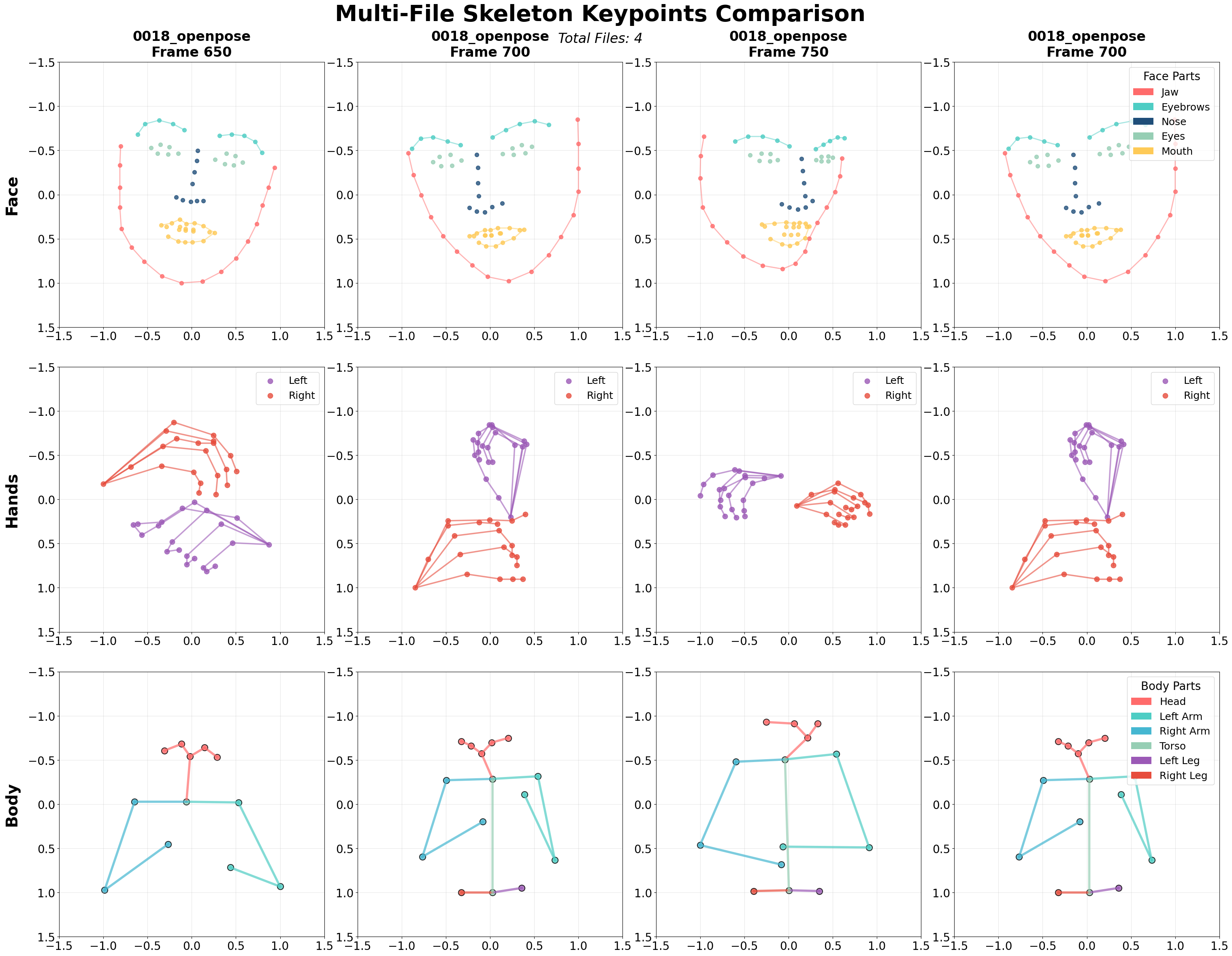}
	\caption{Visualization of Openpose Keypoint Connection}
	\label{fig5}
\end{figure}

\subsection{Data Distribution Visualization}

We analyzed the distribution of the number of frames in the videos. It was found that the number of frames per video sample in the dataset varies greatly. Therefore, truncating or repeating previous frames to align the video length is not feasible when processing the videos. In the end, we adopted a sampling method.
\begin{figure}[h]
	\centering
	\includegraphics[width=1\linewidth]{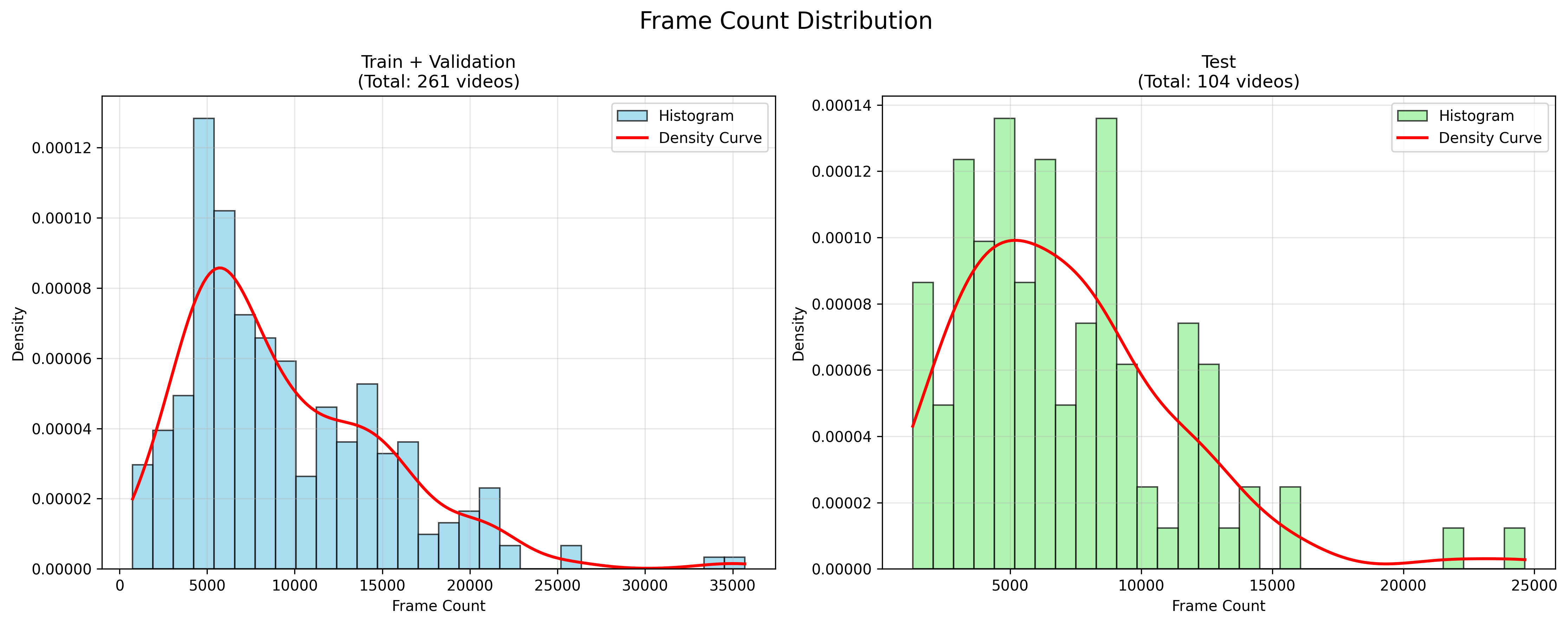}
	\caption{Visualization of frames Distribution}
	\label{fig5}
\end{figure}

\section{Method}

\subsection{Data Processing and Feature Engineering}

We believe that the task of video-based hidden emotion understanding relies both on the model's ability to independently recognize fleeting (lasting less than 0.5 seconds) micro-expressions and micro-gestures in the video, and on its ability to model the changes of all micro-expressions and micro-gestures distributed across thousands of frames, so as to characterize the underlying emotional changes they imply.

Therefore, the key to improving model performance lies in introducing prior knowledge to filter out a large amount of redundant information in the data, thereby efficiently constructing features. This is the design philosophy behind our data processing and feature engineering work.

Because we mainly focus on the expressions and actions of the interviewee in the video, large areas of background in the video are redundant information. Therefore, we use YOLO11x to crop only the rectangular boxes containing human figures in each frame of the video. Considering that Dinov2, trained based on self-supervision, has extremely strong general visual representation capability, this means that the feature vectors it extracts not only have strong discriminative ability on images with subtle differences, but also indicate that these features have good cross-modal alignment capability. At the same time, Dinov2 divides the input image into $37 \times 37$ patches with a width and height of 14 pixels each, and the receptive field size of each patch can just cover the area around several facial and hand keypoints in the cropped image. These characteristics make it naturally suitable for capturing visual representations based on facial muscle and finger movements in a single frame. Therefore, we use the Base-size Dinov2 to encode each frame-cropped image of the video, resulting in an image feature time series with a length equal to the number of frames containing human figures multiplied by 786 dimensions.

We consider that, apart from the background, the athlete's clothes are actually redundant features, but if only the face is retained, the body and gesture information of the athlete will be lost. Therefore, we use the Openpose model to identify 137 human body keypoints and obtain their pixel coordinates in the YOLO11x-cropped images. There are slight differences compared to the keypoint coordinates provided by the dataset; we did not save the confidence scores of the keypoints during processing. At the same time, the pixel coordinates of the keypoints we extract use the upper left corner of the detection box as the origin and are normalized to ratio coordinates relative to the width and height of the detection box (that is, the coordinate values are all between 0 and 1), rather than the absolute pixel positions in the original image. The advantage of this is that it can amplify the positional differences between keypoints within and between frames. At the same time, we believe that for a time series model to model the ultra-long keypoint position sequence and also autonomously capture micro-expression and micro-gesture features, represented by keypoint position changes lasting less than 0.5 seconds, is difficult. Inspired by optical flow algorithms, we explicitly construct features for each frame to represent the positional changes of keypoints, specifically by calculating the keypoint position changes between the current frame and several previous frames as the keypoint features of the current frame. Considering that one second is about 25 frames, we uniformly extract three frames within one second to calculate the keypoint position offsets with the current frame, which avoids micro-expressions and micro-movements disappearing when the interval between frames is too large, and also avoids insignificant keypoint displacements when the interval is too small. The extracted keypoints are labeled as skeleton keypoints 0-24, facial keypoints 25-94, and hand keypoints 95-136. These three groups of keypoints are input to the model separately.

Finally, we note that existing work focuses on improving the models for the visual and human keypoint modalities, while ignoring that the increasingly developed cross-modal large models have the ability to efficiently map video information to other modalities based on the rich cross-modal prior knowledge learned during pre-training. For example, Gemini 2.5 Pro can relatively accurately read action units from FACS, PGCS, and GACS in videos, and analyze the emotional changes of the interviewee in the video based on the emotions corresponding to these action units. However, the model obviously lacks prior knowledge of hidden emotion analysis. We notice that many athletes will make face-scratching actions, which is interpreted by Gemini 2.5 Pro as a self-soothing action to relieve tension, thus determining that the athlete has suffered defeat in the match. But in fact, a more likely reason is that sweat makes airborne foreign matter more likely to adhere to the skin and cause irritation, or the athlete's sweat glands begin to contract after the match, causing itching. If the model cannot associate such prior knowledge, it will make incorrect judgments. We set prior rules in the Prompt and require the model to follow the thinking sequence specified by CoT to first recognize action units, and then, under the guidance of these prior rules, list the evidence for winning and losing instead of directly giving a win or loss judgment. Finally, based on the evidence given by the model, reflect on whether the prior rules are violated, and give confidence scores for winning and losing based on the evidence. This CoT+reflection prompt technique significantly helps the model correct erroneous judgments. After that, for each video, we obtained a reliable text description, which contains the recognizable FACS, PGCS, GACS action units and corresponding emotions, as well as the analysis based on these emotions and the prior rules we provided. In addition, we chose the class with the higher confidence between "win" and "loss" as the pseudo-label for the sample, so our test set also has pseudo-labels.

\subsection{Model Design}

Referring to previous methods, researchers often use CNN-based models to extract features from single video frames and then use LSTM to model feature sequences, or use STGCN to first aggregate spatial features of nodes through GCN and then aggregate temporal features of nodes through CNN. Both methods share common drawbacks: their ability to model long time series such as video frame sequences is insufficient, and they only consider a single modality. According to feature engineering, we transform the video into portrait feature time series modality, keypoint time series modality, and text modality via different pre-trained models.

For the keypoint time series modality, we label keypoints 0-24 as skeleton keypoints, 25-94 as facial keypoints, and 95-136 as hand keypoints. For each group of keypoints, according to the keypoint connection relationship defined by OpenPose, we organize these keypoints as a graph structure in the form of an adjacency matrix.

For each group of keypoint time series inputs, we first use GCN to extract spatial structure features for the graph constructed from the keypoints in each frame and flatten the node features into frame-level features; then, the obtained frame-level feature sequence is input to a Transformer to obtain high-level temporal features of the same length as the input sequence; finally, average pooling is performed along the temporal dimension for all frame features output by the Transformer to compress the temporal information, ultimately obtaining a high-level global feature representation after temporal modeling. However, our design philosophy is still to keep things as simple as possible. From the perspective of node information interaction, organizing node features as a graph structure in the form of an adjacency matrix for GCN input is actually equivalent to concatenating node features and inputting them into a fully connected layer. GCN uses the connection relationships between nodes as prior for feature aggregation, while MLP does not have such prior. If this prior is inaccurate, it may affect the model's performance. Specifically, the keypoint connection prior provided by OpenPose is based on facial organs or bone connections, i.e., there is no connection if the keypoints are not connected by bones or do not belong to the same facial organ. However, micro-expressions often involve facial muscle pulling that causes displacement of keypoints in multiple facial organs, but there is no connection between these keypoints. Similarly, in gestures involving both hands, there are no connections between the keypoints of the left and right hands. This means that the synergistic displacement information between keypoints of different facial organs or between the left and right hands, which should be captured by the model, cannot be effectively transmitted and aggregated in GCN based on the OpenPose structure prior. Thanks to the fully connected structure of the linear layer, any two features can directly interact via the weight matrix without relying on predefined adjacency edges. Therefore, we tried using MLP instead of GCN, and achieved equally good or even more stable results when the amount of data was limited.
We input the three sets of keypoint sequences separately into our proposed GCN-Transformer (or MLP-Transformer), concatenate the three global feature vectors obtained, and compress them to 256 dimensions through a fully connected layer.

Similarly, we input the long portrait feature time series extracted frame by frame by Dinov2 into a Transformer for modeling to obtain high-level temporal features of the same length as the input sequence, and use average pooling to compress the temporal information, thereby obtaining a high-level global feature representation in this modality and compressing the dimension to 256 via a fully connected layer.

For the text description of the video generated by Gemini 2.5 Pro, we directly use BERT-Base to encode it into a feature vector and compress the dimension to 256 via a fully connected layer. Finally, we concatenate the compressed feature vectors of the three modalities and use a fully connected layer with residual connection for modality fusion. The fused features are then provided to the output head, resulting in the complete video hidden emotion understanding model. In fact, we also designed some trimodal Cross Attention modules to try to further improve model performance by enhancing multimodal fusion, but due to the small size of the training set, the Cross Attention module could not fully learn the complex cross-modal interaction patterns, was prone to overfitting, and the loss function converged unstably during training, ultimately failing to outperform the baseline combination of feature concatenation followed by a residual connection linear layer.
\begin{figure}[h]
	\centering
	\includegraphics[width=1\linewidth]{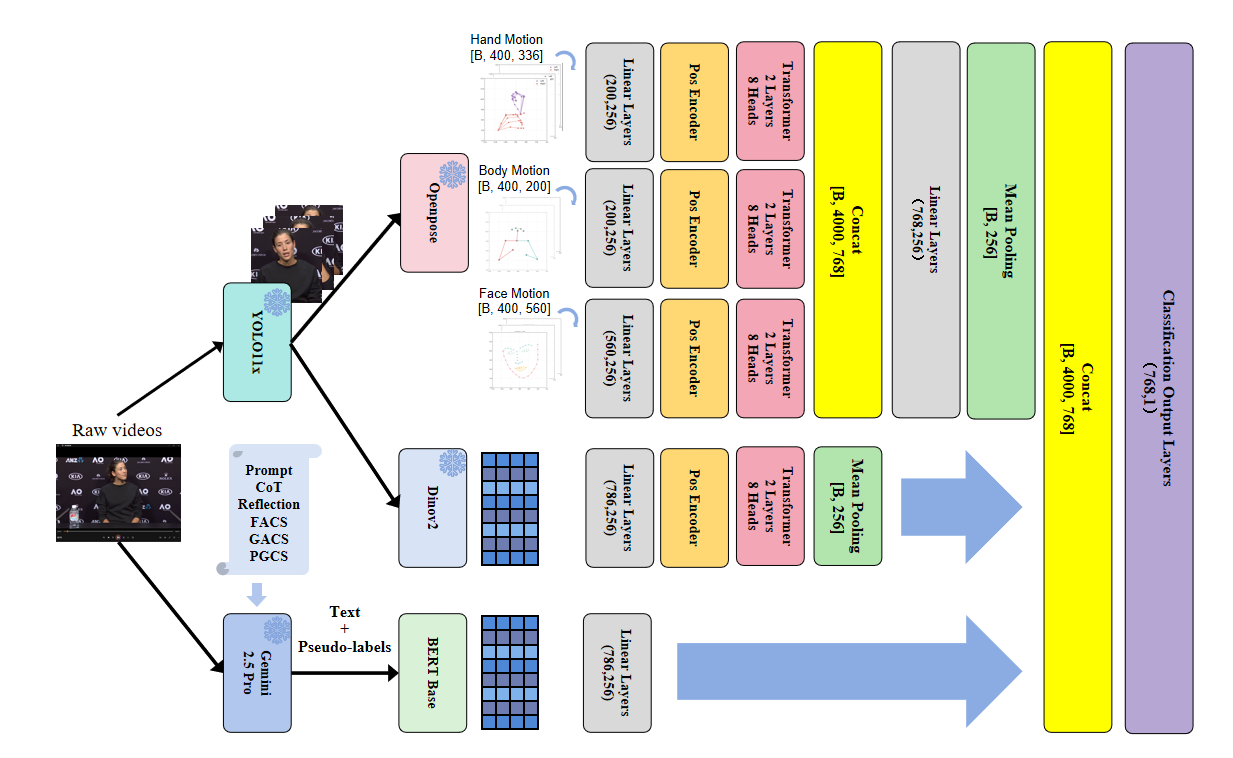}
	\caption{Visualization of Model Structure}
	\label{fig5}
\end{figure}

\subsection{Training Strategy}

We faced two major challenges during training: the scarcity of training data and the inconsistent learning rates of different modalities. Therefore, our training strategy mainly revolves around increasing training data and synchronizing the learning process of different modalities.

Overall, we adopted a two-stage training process. Since the training set is imbalanced in terms of positive and negative samples, we chose Focal Loss as the loss function. In the first stage, we train the networks of each modality separately. It should be noted that YOLO11x, which is used for frame-wise detection and portrait cropping, OpenPose, which is used for keypoint position extraction, and Gemini 2.5 Pro, which converts videos into text descriptions, only provide preprocessing results and are not included in the forward pass of the main network. Therefore, these three are kept frozen during training and do not participate in gradient updates. Thus, for the backbone networks of the keypoint and portrait screenshot feature sequences, we only train the Transformer modules, as well as BERT, which is used as the text description encoder. In this stage, only training data is used as the training set.

It should be pointed out that, except for the text modality, data generation for the keypoint and video modalities both rely on frame sampling (4000 frames for the keypoint modality, 800 frames for the video modality). Therefore, each time the same sample is loaded, differences will occur. We implement data augmentation by repeatedly sampling these modalities and merging the resulting samples, thus improving the generalization ability of the pre-trained model. The practice of obtaining multiple samples from multiple samplings for one sample is also used in inference. In this way, for a test sample, the model can obtain multiple prediction results through multiple sampled inputs and use a voting strategy to obtain the final output, thereby improving the stability and reliability of inference.

In the second stage of training, inspired by the work of Lang et al.\cite{lang2024theoretical}, we designed a weakly supervised training method. In the data preprocessing part, we used Gemini 2.5 Pro to analyze the test set videos under the guidance of our designed Prompt and finally obtained the confidences for the interviewees' victory and failure. We generated pseudo-labels for each test set sample based on the confidences, achieving a test set accuracy of 64.4\%. We combined the original training set and the pseudo-labeled test set to form a new weakly supervised learning dataset. Next, we loaded the best-performing pre-trained backbone network for each modality back into the complete video hidden emotion understanding model. Then, using this weakly supervised learning dataset, we fine-tuned the complete model (with the pre-trained backbones of each modality loaded) with a small learning rate to obtain the final model.

It is worth mentioning that during the first stage of training, we also tried to add a small-weight contrastive learning loss in an attempt to enhance the discriminative ability of the pre-trained model. In the second stage of training, we tried to iteratively refine the pseudo-labels for the test set via semi-supervised learning after completing weakly supervised learning by using model-inferred new test set labels. However, after multiple repeated experiments, we did not obtain sufficient evidence that these two improvements could stably improve model performance. Therefore, following the philosophy of a simplified training strategy, we did not introduce them as part of the formal training pipeline.

In addition, we omitted some practical enhancement tricks used in the competition, such as quantile thresholding and model ensembling. The effectiveness of quantile thresholding depends on the strong assumption that the training set and test set are from the same distribution, but actual data seem unlikely to meet this assumption. Therefore, the quantile threshold for the test set in the competition was obtained by multiple submission attempts. As for model ensembling, as a recognized method of trading time and space for a small improvement in accuracy, it cannot be regarded as an actual contribution.

\section{Experiments}
In order to demonstrate the effectiveness of our proposed method and the chosen model architecture, we conducted experiments using the accuracy of the model on the test set as the evaluation metric. Since the test set labels are not available to us, the model scores are calculated by the official platform. Likewise, we are also unable to obtain AUC or F1 Score.
Our comparative experiments are organized into four parts: (1) selecting the optimal keypoint feature extraction network, (2) verifying the performance gain from explicit offset features, (3) using XCLIP as a baseline to evaluate the improvement of frame-wise feature extraction via Dinov2 and temporal modeling with a Transformer, and (4) analyzing the impact of modality-specific pretraining and two-stage weakly supervised learning on model accuracy.

For spatial modeling of keypoints, we compared four classic graph neural network architectures—Graph Convolutional Network (GCN), Graph Attention Network (GAT) and Graph Isomorphism Network (GIN) as well as a Multilayer Perceptron (MLP) baseline that directly processes keypoint coordinates. The results are as follows:
\FloatBarrier
\begin{table}[H]
    \centering
    \begin{tabular}{l|c}
        \hline
        Network & ACC \\
        \hline
        GCN & 67.31\% \\
        GIN  & 56.73\% \\
        GAT  & 62.50\% \\
        MLP & 66.35\% \\
        MLP + offset features & 68.27\% \\
        GCN + offset features & 68.27\% \\
        \hline
    \end{tabular}
    \caption{Classification accuracy of different keypoint feature extraction networks}
\end{table}
\FloatBarrier
The results show that while graph neural networks excel in various graph-structured tasks, in our scenario, MLP achieves the best classification performance with simpler and more efficient feature mapping (accuracy comparable to GCN but with faster training).

We further used the large-scale video pre-trained model XCLIP as a baseline to compare with our long-sequence feature extraction pipeline based on Dinov2+Transformer. The results are as follows:
\FloatBarrier
\begin{table}[H]
    \centering
    \begin{tabular}{l|c}
        \hline
        Network & ACC \\
        \hline
        Dinov2 Base & 68.27\% \\
        X-CLIP Base (16 Frames)& 67.31\% \\
        \hline
    \end{tabular}
    \caption{Classification accuracy of different video feature extraction models}
\end{table}
\FloatBarrier
Although pretraining enables XCLIP to better model global context based on sampled video frames and extract global features, it samples only 16 frames per video, making it difficult to capture fleeting micro-expressions. As a result, it relies mainly on explicit expressions and actions, thereby missing substantial fine-grained information crucial for our task. In contrast, our method performs dense frame sampling and models long temporal sequences with a Transformer, greatly enhancing the model’s ability to perceive and capture subtle dynamic cues.

Finally, we compared the impact of weakly supervised learning with pseudo-labels generated by Gemini 2.5 Pro, direct multimodal joint training, modality-specific pretraining, and introducing contrastive loss during pretraining. The results are summarized below:
\FloatBarrier
\begin{table}[H]
    \centering
    \begin{tabular}{l|c}
        \hline
        Training Strategy & ACC \\
        \hline
        MLP & 63.46\% \\
        \makecell[l]{MLP\\+ weak supervision} & 66.35\% \\
        \makecell[l]{MLP\\+ weak supervision\\+ offset features} & 68.27\% \\
        \makecell[l]{MLP\\+ offset features\\+ pretraining\\+ weak supervision} & 69.23\% \\
        \makecell[l]{MLP\\+ offset features\\+ contrastive\\+ pretraining\\+ weak supervision} & 68.27\% \\
        \hline
    \end{tabular}
    \caption{Accuracy comparison of different training strategies}
\end{table}
\FloatBarrier
The results demonstrate that modality-specific pretraining followed by overall weakly supervised fine-tuning yields the greatest improvement in model accuracy.

\section{Conclusion}
\subsection{Limitations and Reflections}%
\label{subsec:limitations}

\begin{enumerate}
  \item \textbf{Inaccurate hand keypoints.}  
  Inspection of OpenPose outputs reveals that the detection quality of hand keypoints is extremely poor, introducing substantial noise into the hand–keypoint modality. Future work will employ a state\-of\-the\-art hand keypoint detector to re\-estimate these landmarks.

  \item \textbf{Lack of a theoretical framework for learning with noisy labels.}  
  Our weak\-supervision strategy mixes label\-noisy data with the official training set, yet we provide neither a formal hypothesis nor an empirical study of its generalisation error bounds. Specifically, we have not analysed why performance may degrade once the overall noise rate exceeds a certain threshold, nor identified the optimal noise\-rate window in which weak supervision is beneficial. Establishing such a theoretical framework is one of our next goals.

  \item \textbf{Unstable gains from semi\-supervised learning.}  
  Previous experiments did not show consistent improvements from semi\-supervised learning, potentially because we failed to filter high\-confidence samples before re\-feeding them into the next training round. We will explore more effective semi\-supervised strategies to integrate into the training pipeline and further enhance performance.

  \item \textbf{Limited fine\-tuning and hyperparameters selection.}  
  Owing to data scarcity, we did not fine\-tune DINOv2 or use more powerful text encoders, which caps overall performance. Moreover, we did not reserve a validation set to detect overfitting or to determine post\-processing parameters (e.g.\ quantile thresholds) automatically; key hyperparameters such as training epochs and decision thresholds were selected via leaderboard feedback. This assumes no distribution shift between training and test sets—a questionable premise in practice.
\end{enumerate}

\subsection{Future Work}%
\label{subsec:future}

\begin{enumerate}
  \item \textbf{Leveraging 4D facial analysis.}  
  Recent advances in 4D facial analysis—capturing the temporal evolution of facial expressions in 3D space—have greatly boosted micro\-expression recognition. Concurrently, \emph{GAGAvatar}, built upon a 3D Morphable Model (3DMM) \cite{li2017learning}, can reconstruct high\-fidelity, animatable 3D Gaussian representations from monocular video and returns 5{,}023 topology\-consistent key vertices with learnable feature vectors. We plan to adopt GAGAvatar\cite{chu2024generalizable} as an intermediate representation generator, converting raw video into a set of animatable 3D Gaussian parameters along with expression and pose coefficients, thereby supplying micro\-expression recognition with fine\-grained and cross\-view\-consistent 4D geometry and appearance descriptors.

  \item \textbf{Binding 2D keypoints to semantic patches.}  
  Inspired by 3DMM’s one\-to\-one binding between vertex geometry and learnable feature vectors, we aim to establish a similar correspondence between 2D keypoints and visual features. Concretely, we will first detect 137 keypoints in each frame using OpenPose and map their pixel coordinates to the corresponding image patches produced by DINOv2. The high\-level semantic feature vector of each patch serves as the keypoint’s visual embedding, while the normalised pixel coordinates and their offsets relative to several previous frames jointly encode positional information. By binding \emph{semantic, spatial,} and \emph{temporal} cues at the keypoint level, we compress the high\-dimensional video modality into a sparse, structured feature set that offers more discriminative input for subsequent temporal modelling.
\end{enumerate}

\subsection{Conclusion}%

In line with the philosophy of simplicity and efficiency, this paper proposes a trimodal weakly supervised framework for video-based hidden emotion understanding on the iMiGUE dataset. The methodological pipeline is as follows: First, YOLO 11x is utilized to precisely crop human regions in each frame, and Dinov2-Base is employed to encode features of the cropped portraits, forming an ultra-long sequence of image features. Second, OpenPose is used to extract 25 skeletal, 70 facial, and 42 hand keypoints, and cross-frame displacements are computed; the three sets of keypoint sequences are structured as graphs and modeled via a Transformer. Meanwhile, combining Chain-of-Thought (CoT) and Reflection prompts, both the video and prior rules are fed into Gemini 2.5 Pro to generate action units and the reasoning text for victory/failure inference, which are then encoded by BERT to supplement the textual modality. The outputs from the three Transformer branches are concatenated and fused via a fully connected layer before being passed to the classification head.

To address label scarcity, pseudo-labels generated by Gemini are further incorporated into the training set to construct a weakly supervised dataset for joint fine-tuning. Comparative experiments show that this framework achieves state-of-the-art performance on hidden emotion recognition accuracy, setting a new benchmark on iMiGUE. Ablation studies reveal that replacing the GCN backbone in the human keypoint branch with an MLP can maintain or even improve performance while reducing computational cost, confirming that the spatiotemporal relationships in this task can be sufficiently modeled by an MLP+Transformer architecture without explicit graph convolutions.
\begin{acknowledgments}
We would like to thank the MiGA-IJCAI Workshop organizing team for hosting such a high-quality challenge and for providing the valuable iMiGUE dataset.
We would like to thank all members of the Backpacker team for their equal contributions during the competition. Haixu Liu proposed the key improvement of using Gemini 2.5 Pro to generate pseudo-labels and conducted multimodal weakly supervised training, as well as revised the paper. Yufei Wang completed all the code, including model design, training strategy, ablation experiments, and data visualization. Tiangxiang Xu collaborated with Yufei Wang to write the manuscript. Chuancheng Shi was responsible for polishing the English writing, and Hongsheng Xing created the model schematic diagrams.
\end{acknowledgments}


\section*{Declaration on Generative AI}
 During the preparation of this work, the authors used ChatGPT-4.5 and ChatGPT-o3 in order to: Text Translation and Formatting assistance. After using these tools/services, the authors reviewed and edited the content as needed and take full responsibility for the publication’s content. 
\bibliography{sample-ceur}



\end{document}